\documentclass[11pt]{article}

\usepackage[preprint]{acl}

\usepackage{times}
\usepackage{latexsym}
\usepackage{amsmath}
\usepackage{bbm}
\usepackage{multirow}
\usepackage[T1]{fontenc}
\usepackage[utf8]{inputenc}
\usepackage{float}
\usepackage{microtype}
\usepackage{enumitem}
\IfFileExists{tcolorbox.sty}{\usepackage[most]{tcolorbox}}{}
\IfFileExists{inconsolata.sty}{\usepackage{inconsolata}}{}
\usepackage{xcolor}
\usepackage{colortbl}
\usepackage{graphicx}
\usepackage{fontawesome5} 

\usepackage{booktabs}
\usepackage{tabularx}
\newcolumntype{R}{>{\raggedleft\arraybackslash}X}
\newcolumntype{L}{>{\raggedright\arraybackslash}X}
\newcolumntype{C}{>{\centering\arraybackslash}X}

\IfFileExists{tcolorbox.sty}{
  \newtcolorbox{finding}{
    enhanced, breakable,
    colback=black!4, colframe=black!55,
    boxrule=0.5pt, arc=2pt,
    left=6pt, right=6pt, top=4pt, bottom=4pt,
    boxsep=2pt,
  }
}{
  
  \newenvironment{tcolorbox}[1][]{\begin{quote}\small}{\end{quote}}
}

\title{Thinking Hard, Not Smart: \\Reasoning Models Fail to Ration Test-Time Compute Across Questions}

\author{%
  Chenrui Fan\thanks{Equal contribution}$^1$, Yize Cheng\footnotemark[1]$^1$, Ming Li$^1$, Yongyuan Liang$^1$, Tianyi Zhou$^2$, Soheil Feizi$^1$  \\
  $^1$University of Maryland, College Park~~~~~~$^2$MBZUAI, UAE\\
  \texttt{\{cfan42, yzcheng, minglii, cheryunl, sfeizi\}@umd.edu, tianyi.zhou@mbzuai.ac.ae} \\
    \faGithub~Project: \url{https://github.com/Fcr09/thinking-hard-not-smart}
}

\begin{document}
\maketitle
\begin{abstract}
Reasoning language models increasingly use test-time compute to improve performance, but existing evaluations typically study this compute one question at a time. 
Yet when multiple problems share an end-to-end cost or latency constraint, models must decide how to divide limited inference compute among them.
We introduce an exam-style evaluation framework for studying this setting, in which a model must distribute one shared token budget across questions with different difficulty and point values to maximize its total score. 
Across several open and frontier reasoning models, we find that models fail to allocate a shared budget strategically across questions of varying difficulties and values.
Models behave largely as greedy sequential solvers: they prioritize questions by presentation order, front-load effort on early questions, and remain insensitive to value, with these tendencies becoming more pronounced as the number of questions grows. Explicit planning prompts spread compute more evenly but do not produce value- or difficulty-aware prioritization. The same behavioral pattern extends from mathematical to code reasoning. These findings establish global budget allocation as a distinct capability that is not captured by conventional per-question evaluation and remains a challenge for current reasoning models.
\end{abstract}

\section{Introduction}

\begin{figure}[t]
    \centering
    \includegraphics[width=1\linewidth]{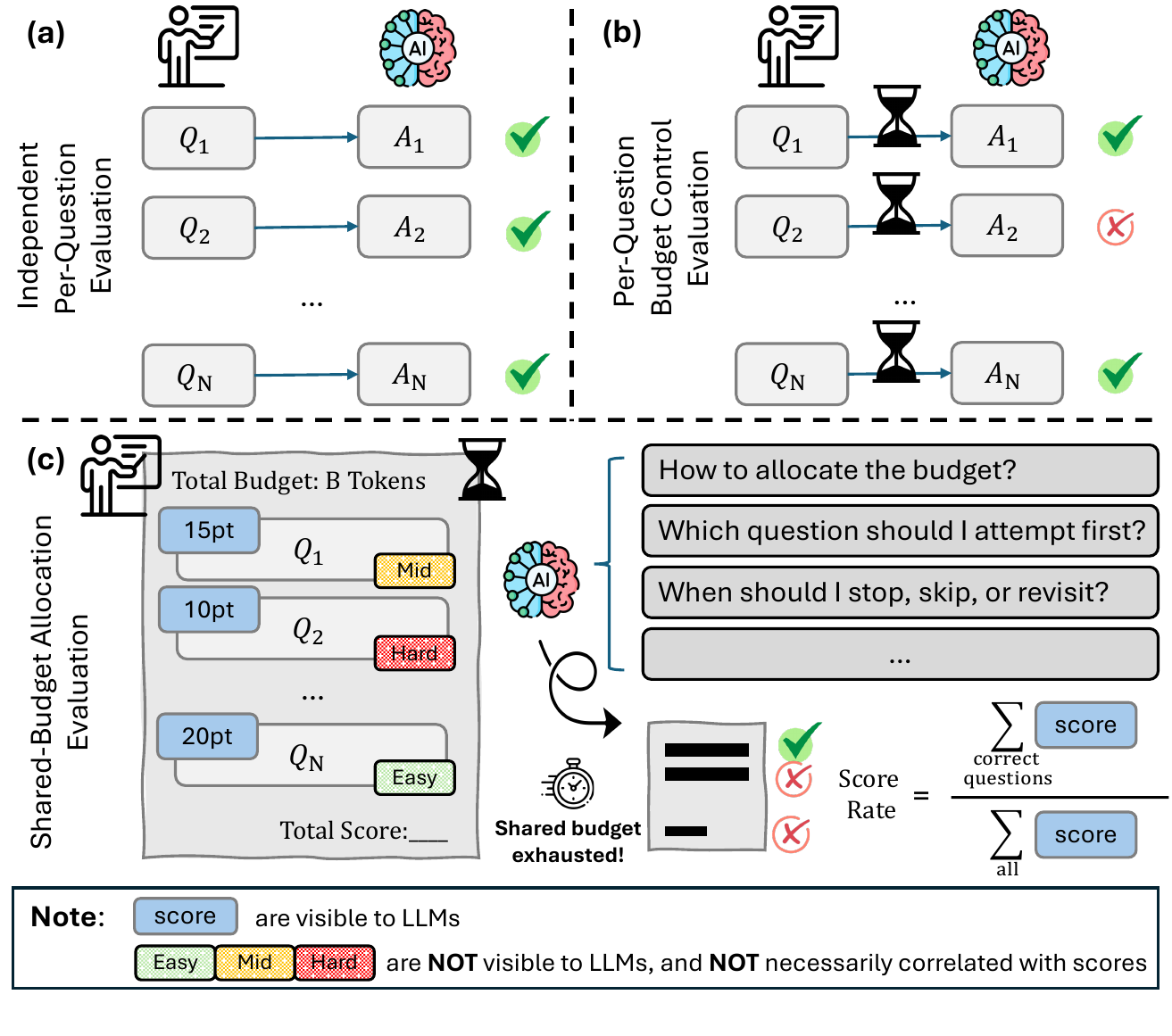}
    \caption{\textbf{Three evaluation regimes.} (a) Standard: one question with an unrestricted budget. (b) A separate budget cap for each question. (c) \textbf{Our setting:} $N$ scored questions compete for one global budget, which tests cross-question allocation.}
    \label{fig:teaser}
    \vspace{-3mm}
\end{figure}

Reasoning models have become stronger by learning to spend more inference-time computation on difficult problems~\cite{openai2024o1,Guo_2025,snell2024scalingllmtesttimecompute,muennighoff2025s1}. Yet more reasoning is not always useful, and models can overthink even a single problem~\cite{chen2025think23overthinkingo1like,ma2025reasoningmodelseffectivethinking,fan2025missingpremiseexacerbatesoverthinking}. When several problems compete for a finite budget, the decision becomes harder because continuing one problem leaves less computation for the others. A model must know not only how to solve a problem, but also which problems are worth attempting, when to give up, and when to return. Conventional one-question-at-a-time evaluation conceals this ability because each problem receives its own budget and creates no tradeoff across problems.

An exam-style evaluation offers a controlled probe for this: multiple questions with visible point values compete for one total budget, and total score supplies a concrete objective. This resembles a knapsack problem~\cite{kellerer2004knapsack} in which the budget is the capacity, question scores are values, and model-specific solution costs are weights. Because costs and success probabilities are not explicitly provided, a strategic model must estimate value relative to cost, revise that judgment while reasoning, and abandon attempts that no longer justify further effort. We ask whether reasoning models exhibit this form of metacognitive control when allocating computation across questions.

Existing work controls inference effort for one problem at a time through explicit limits, adaptive computation, or difficulty-conditioned budgets~\cite{aggarwal2025l1controllinglongreasoning,wang2025adareasoneradaptivereasoningenables,wu2025armadaptivereasoningmodel,wen2025budgetthinkerempoweringbudgetawarellm,han2025tokenbudgetawarellmreasoning}. Batch prompting groups multiple questions into one request to amortize shared instructions and reduce inference cost~\cite{cheng2023batch}, with related work studying multi-problem evaluation and composed instructions~\cite{wang2025evaluating,li2025mosaic}. REST~\cite{pan2025rest} more directly stress-tests reasoning models by presenting multiple problems at once and studying their degradation under multi-context pressure. These settings do not center how realized reasoning effort responds to question value, cost, and presentation under one explicit budget. Concurrent work, TRIAGE~\cite{nazi2026triage}, evaluates an allocation plan committed before execution; we instead study the allocation that emerges while a model jointly executes the questions, with freedom to reorder, defer, revisit, or abandon them.

We operationalize this probe through the exam-style evaluation illustrated in Figure~\ref{fig:teaser}. Each exam leaves the model free to decide which questions to attempt, in what order, and with how much effort. We construct matched exams from Omni-MATH~\cite{gao2024omnimathuniversalolympiadlevel} and systematically vary the length, presentation order, and point values of the same questions. These controlled variants let us
distinguish a simple sensitivity to presentation position from  genuine sensitivity to question value and difficulty.
We also compare direct solving with planning instructions to test whether an explicit opportunity to allocate the budget improves this behavior.
% From each joint reasoning trace, we recover the model's realized effort and solving order. We then compare this allocation with independent high-budget attempts on each question, using their correctness and token usage as model-specific empirical proxies for solvability and solution cost. 
Our study covers five locally deployed open-weight models and two DeepSeek-V4 API models. The same findings also generalize to a code domain on CRUXEval-O~\cite{pmlr-v235-gu24c}.

\paragraph{Key Findings.}
\begin{itemize}[leftmargin=*,itemsep=2pt,topsep=2pt]
    % \item \textbf{Allocation follows position rather than value.} Models largely solve questions in presentation order, spend progressively less on later questions, and respond little to stated point values.
    \item \textbf{Models allocate compute sequentially rather than strategically.} Models largely solve questions in presentation order, spend progressively less on later questions, and respond little to stated point values. Their allocation is therefore governed more by which question appears next than by which question is most worth attempting.
    % Difficulty affects effort mainly after a question is attempted, which suggests reactive persistence rather than prospective prioritization.
    \item \textbf{Budget pressure magnifies the failure.}
    Averaged across all models, as the exam length $N$ increases, the correlation between solving order and presentation position strengthens, while the coverage of problems on which models expend substantial effort decreases monotonically.
    % When exam length $N=20$ (20 questions), locally deployed open-weight models perform substantive work on only $40\%$ of questions and allocate no reasoning tokens to $51\%$. Separately, 
    % when $N=10$, the problems that the models spent 
    % in our $N=10$ density analysis, questions not solved by the same model in the independent high-budget reference still consume $32\%$ of its shared reasoning budget.
    \item \textbf{Planning changes spread, not priorities.} Planning instructions improve coverage but do not induce value-aware allocation. Reordering and repricing the same questions produce little strategic adaptation. The same position-driven pattern generalizes to code reasoning.
\end{itemize}

\paragraph{Contributions.}
We design a controlled framework for studying shared-budget reasoning, a trace-based analysis of realized effort and solving order, and broad evidence that object-level reasoning ability does not ensure strategic control across questions. Current models know how to think hard about the question in front of them, but not how to decide which question is worth thinking about.

\section{Related Work}

\paragraph{Per-problem test-time compute control.}
Scaling inference-time computation can improve reasoning~\citep{snell2024scalingllmtesttimecompute,Guo_2025,muennighoff2025s1}, but can also lead to overthinking~\cite{chen2025think23overthinkingo1like}. Existing methods improve efficiency through length control~\citep{aggarwal2025l1controllinglongreasoning,hou2025thinkprunepruninglongchainofthought,li2025steering,li2025makes}, adaptive effort~\citep{wang2025adareasoneradaptivereasoningenables,wu2025armadaptivereasoningmodel}, and difficulty-conditioned budgets~\citep{han2025tokenbudgetawarellmreasoning,wen2025budgetthinkerempoweringbudgetawarellm}. Related work also studies budget-aware evaluation~\citep{wang-etal-2024-reasoning-token}, anytime reasoning~\citep{zhang-etal-2026-budget}, and the structure and metacognitive control of reasoning trajectories~\citep{li-etal-2025-understanding,li-etal-2026-schoenfelds,ma2026cot2metabudgetedmetacognitivecontrol}.
% \textcolor{red}{[CITE: Reasoning in Token Economies; Budget-Aware Anytime Reasoning with LLM-Synthesized Preference Data; CoT2-Meta]}. 
These approaches decide how much computation to spend on a given problem. We instead study the opportunity cost created when several questions compete for the same budget.

\paragraph{Multi-question prompting and position sensitivity.}
Batch prompting groups multiple questions into one request to amortize shared instructions and reduce inference cost~\citep{cheng2023batch}, while other work evaluates models on multiple problems or composed instructions~\citep{wang2025evaluating,li2025mosaic}. REST presents several reasoning problems simultaneously to study multi-context degradation and contextual priority allocation~\citep{pan2025rest}. A separate literature~\citep{chen2024premiseordermattersreasoning,schilcher-etal-2025-characterizing} shows that model behavior can be sensitive to the ordering of prompt elements or reasoning premises.
% \textcolor{red}{[CITE: Premise Order Matters in Reasoning with Large Language Models; Characterizing Positional Bias in Large Language Models]}. 
These studies establish multi-problem interference and order sensitivity, but do not center how realized reasoning effort responds to visible question values and model-specific costs under one explicit global budget.

\paragraph{Global allocation and metareasoning.}
Rational metareasoning treats computation itself as a decision, using its expected value to determine which reasoning operation is worth performing~\citep{RUSSELL1991361, desabbata2025rationalmetareasoninglargelanguage}.
% \textcolor{red}{[CITE: Principles of Metareasoning; Definition and Complexity of Some Basic Metareasoning Problems; Rational Metareasoning for Large Language Models]}. 
Recent work~\citep{zhai2026adaptivetesttimecomputeallocation} begins to allocate test-time compute across inputs using learned per-instance budget policies.
% \textcolor{red}{[CITE: Adaptive Test-Time Compute Allocation for Reasoning LLMs via Constrained Policy Optimization]}. 
ROI-Reasoning~\citep{zhao2026roi} trains models for knapsack-style solve-or-skip allocation under a global token cap, but retains a fixed processing order. 
Concurrently, TRIAGE~\citep{nazi2026triage} evaluates the quality of an upfront plan, in contrast, we diagnose the allocation that emerges jointly during solving, where the model can reorder, defer, or abandon questions mid-trace.
% Concurrent TRIAGE evaluates a prospective plan for question selection, ordering, and per-problem budgets~\citep{nazi2026triage}. In contrast, we diagnose the allocation that emerges during joint execution, where the model may reorder, defer, revisit, or abandon questions as it reasons and where question cost and solvability are not provided in advance.

% \input{sec/setup}
\section{Shared-Budget Multi-Question Reasoning}
\label{sec:setup}

We evaluate whether reasoning models can distribute a finite inference budget across multiple competing questions. 

\subsection{Task Formulation}
\label{sec:task}

An exam is
\begin{equation}
    E=\{(q_i,v_i)\}_{i=1}^{N},
\end{equation}
where $q_i$ is a question and $v_i$ is its visible point value. In each exam, the model sees all questions and their point values at once, receives a shared budget of $B$ reasoning tokens, and aims to maximize score rate:
\vspace{-0.2cm}
\begin{equation}
    \frac{1}{\sum_{i=1}^{N}v_i}\sum_{i=1}^{N} v_i
    \mathbbm{1}[\hat{a}_i=a_i]. \label{eq:score_rate}
\end{equation}

In our setting, inference uses two phases. First, the model produces one reasoning trace for the entire exam, capped at $B$ generated tokens. The prompt does not require any particular solving order or budget split. Second, after the reasoning phase ends, we ask for the final answer in a separate follow-up turn, using the existing reasoning trace as conversation history. This phase is used only for answer extraction and does not count toward the shared budget.

% For mathematics, we set $B=20{,}000$. The limit is applied using each model's own tokenizer and counts only tokens generated during the reasoning phase.

\subsection{Dataset and Exam Construction}
\label{sec:data}

Our primary experiments use Omni-MATH \citep{gao2024omnimathuniversalolympiadlevel}. We uniformly sample problems whose benchmark difficulty label is at most 5 and construct exams with $N\in\{5,10,20\}$. Difficulty labels are used for experimental construction and analysis but are not shown to the model.
For each value of $N$, we sample 50 base exams, each consisting of a fixed set of questions. The same exams are reused across scoring schemes, question orderings, prompting strategies, and models, so comparisons differ only in the factor being varied.

\subsection{Models and Decoding}
\label{sec:models}

We evaluate five locally hosted open-weight reasoning models: DeepSeek-R1-Distill-Qwen-7B/14B (DQ-7/14) \cite{guo2025deepseek} and Qwen3-8B/14B/32B (QW-8/14/32) \cite{yang2025qwen3}, served via vllm~\cite{kwon2025vllm}. We also evaluate the preview API versions of DeepSeek-V4 Flash and Pro (DSV4-F/P)~\cite{deepseekai2026deepseekv4highlyefficientmilliontoken}.

For locally hosted models, reasoning uses temperature $0.6$, top-$p=0.95$, and top-$k=20$. The API models are run with their available default controls. Answer extraction uses greedy decoding for all models. Mathematical answers are parsed from \texttt{\textbackslash boxed\{\}} outputs and evaluated with an LLM-based judge described in Appendix~\ref{append:Prompts}.

\subsection{Experimental Factors}
\label{sec:factors}

We vary score assignment, question order, and prompting strategy.

\paragraph{Scoring scheme.}
We consider four schemes:

\begin{itemize}[leftmargin=12pt,itemsep=-2pt,topsep=1pt]
    \item \textbf{Fixed:} every question is worth 10 points.
    \item \textbf{Random:} each question receives an integer score from 1 to 15, independently of difficulty and position.
    \item \textbf{Aligned:} harder questions receive more points.
    \item \textbf{Reversed:} easier questions receive more points.
\end{itemize}

For aligned scoring, difficulties are normalized within each exam and mapped to the integer range $[1,15]$:
\begin{equation}
    v_i=
    \left\lfloor
    1+14\frac{d_i-d_{\min}}{d_{\max}-d_{\min}}
    \right\rfloor,
\end{equation}
where $d_i$ is the difficulty for question $i$. For reversed scoring,
\begin{equation}
    v_i=
    \left\lfloor
    15-14\frac{d_i-d_{\min}}{d_{\max}-d_{\min}}
    \right\rfloor.
\end{equation}
If all questions have the same difficulty, each receives 10 points.

\paragraph{Question order.}
Each exam is presented in one of three orders: random (\texttt{rand}), ascending difficulty (\texttt{asc}), or decreasing difficulty (\texttt{dsc}). Random order separates position from difficulty, while the sorted conditions test whether models can depart from the presented sequence when early questions are especially easy or difficult.

\paragraph{Prompting strategy.}
The baseline prompt states the shared budget and the goal of maximizing total score. The explicit-planning condition additionally tells the model that questions may differ in difficulty and reasoning cost and asks it to plan its allocation wisely. Other prompts involving skipping and rechecking are reported in Appendix~\ref{append:Prompts}.

\subsection{Attributing Reasoning to Questions}
\label{sec:allocation}

The reasoning phase produces a single free-form trace for the entire exam. As the questions are presented with identifiers \texttt{Q1}, \texttt{Q2}, \ldots, we use these markers to recover a per-question view of the trace: text between two consecutive question markers is attributed to the earlier question as an approximation.

This attribution allows us to study both how much reasoning each question receives and when it is considered. Because a brief mention need not correspond to a substantive attempt, later analyses also distinguish questions that receive meaningful work from those that are only referenced in passing. We introduce the corresponding measures alongside their results in \S~\ref{sec:results}.

\section{Can Reasoning Models Ration Shared Compute?}
\label{sec:results}

We first characterize the allocation policy that emerges by default by examining which signals govern it (\S\ref{sec:allocation_results}) and whether its budget reaches the questions worth attempting (\S\ref{sec:right_questions}). We then test whether this policy adapts by perturbing different variables. We instruct the model to plan its allocation (\S\ref{sec:planning}) and vary the order and point values of the same questions (\S\ref{sec:order_price}). Results on code reasoning close the section (\S\ref{sec:crux_results}).

\subsection{What Governs the Allocation of Reasoning?}
\label{sec:allocation_results}

To examine how the shared budget is distributed across questions, we study two complementary aspects of allocation:

\begin{itemize}[leftmargin=12pt,itemsep=-2pt,topsep=1pt]
    \item \textbf{Token effort:} how much reasoning a question receives;
    \item \textbf{Solving order:} when the model substantively works on it.
\end{itemize}

Both are computed from the marker segmentation of \S~\ref{sec:allocation}. Let $S_i$ be the set of reasoning segments attributed to question $i$, where each segment $s$ is a sequence of tokens of length $\tau_s$ that begins at position $p_s$. The token effort $t_i$ on question $i$ is the total attributed length across different reasoning segments of the question,
\begin{equation}
    t_i = \sum_{s\in S_i}\tau_s ,
    \label{eq:token_effort}
\end{equation}
Since a question can be mentioned in the plan but never receive substantive reasoning effort, we also define the \emph{work set}
\begin{equation}
    W = \bigl\{\, i \;|\; t_i \ge 200 \ \text{ or } \ |S_i| \ge 2 \,\bigr\},
    \label{eq:work_set}
\end{equation}
which separates substantive attempts from brief references.

Solving order is then defined on the work set. We locate question $i$ by the token-weighted centroid of its segments,
\begin{equation}
    c_i = \frac{\sum_{s\in S_i}\tau_s\,p_s}{\sum_{s\in S_i}\tau_s} ,
    \label{eq:centroid}
\end{equation}
and rank the questions in $W$ by ascending $c_i$. The resulting rank is the solving order. Computing centroids of reasoning segments prevents an early mention from displacing a question that is mostly worked on later, and restricting to $W$ prevents a question that is merely enumerated in an opening pass from being ranked ahead of questions that actually received effort.

\begin{figure*}[t]
    \centering
    \includegraphics[width=\textwidth]{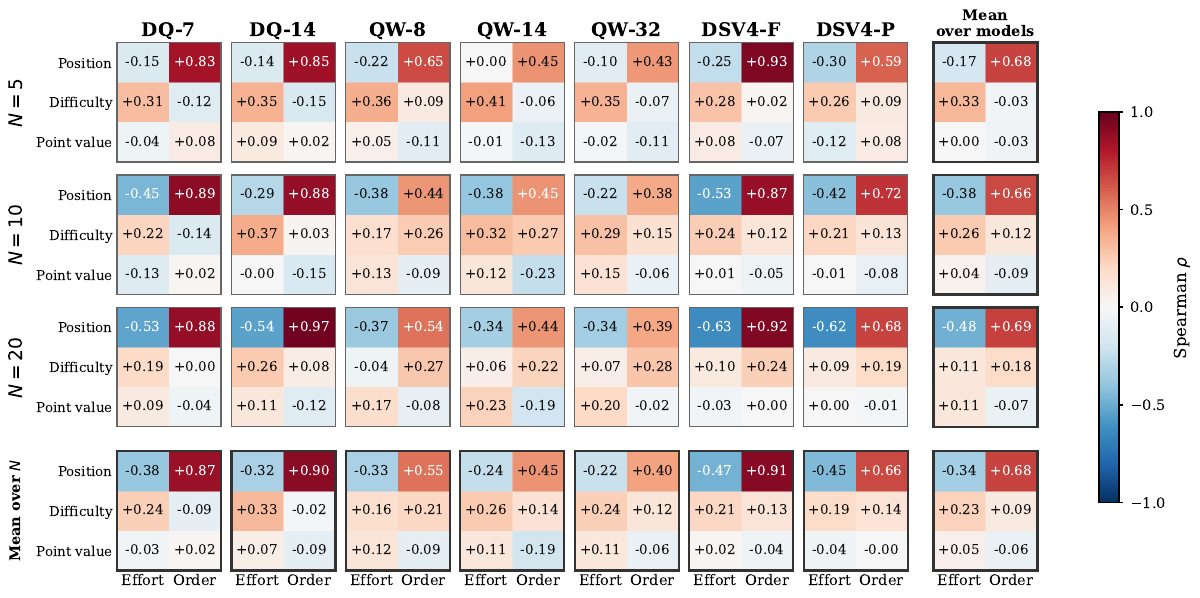}
    \caption{
        Relationships between allocation behavior and presentation
        position, difficulty, and point value under the baseline prompt.
        Position and difficulty are measured using partial Spearman
        correlations under fixed scoring and random question order (first 2 rows of each heat map);
        point value uses ordinary Spearman correlation under random
        scoring (3rd row of each heat map).
        % Columns distinguish token effort from token-centroid
        % solving order. 
        Marginal panels average over models (right), exam
        lengths (bottom), and both (corner).
        DQ denotes DeepSeek-R1-Distill-Qwen, QW denotes Qwen3, and
        DSV4-F/P denote DeepSeek-V4 Flash/Pro. Numeric suffixes indicate
        parameter counts in billions.
        % Blue denotes negative and
        % red positive correlation.
    }
    \label{fig:allocation_correlations}
    \vspace{-3mm}
\end{figure*}

We next ask which information governs these two allocation decisions. For
each question, we consider three candidate signals: its \textbf{presentation
position $\mathbf{\pi_i}$}, its \textbf{difficulty $\mathbf{d_i}$}, and its \textbf{assigned point value
$\mathbf{v_i}$}. We examine whether each signal predicts either the amount of effort the question receives or the order in the trace where it is worked on.

Difficulty and the presented position require some care because they can be correlated within a particular collection of exams. For example, even when questions are randomly ordered, a finite sample may happen to place more difficult questions toward the beginning or end. An ordinary correlation between token effort and position could then partly reflect difficulty, rather than position itself; conversely, an apparent difficulty effect could arise because difficult questions happened to be presented in earlier or later positions.
We address this confound in the fixed-scoring, random-order condition,
where point values are constant and position is independent of difficulty
in expectation. 

To further address remaining position--difficulty correlation that may occur in some samples, we use \emph{partial Spearman
correlations}, which
measures the association between two variables after
controlling for a third. For example, the relationship between
effort and position when controlling difficulty is
\begin{equation}
    \rho_{t,\pi\mid d}
    =
    \frac{
        \rho_{t,\pi}-\rho_{t,d}\rho_{\pi,d}
    }{
        \sqrt{
            \left(1-\rho_{t,d}^{2}\right)
            \left(1-\rho_{\pi,d}^{2}\right)
        }
    },
    \label{eq:partial}
\end{equation}
where each term on the right is an Spearman correlation. We analogously compute $\rho_{t,d\mid\pi}$ to measure the relationship between effort and difficulty after controlling for position. The same procedure is applied to solving order, yielding $\rho_{o,\pi\mid d}$ and $\rho_{o,d\mid\pi}$.

Point value is analyzed separately under random scoring. In this
condition, scores are assigned independently of both difficulty and
position, so neither variable provides a systematic alternative
explanation for an effort--value or order--value relationship. We
therefore report ordinary Spearman correlations with point value rather
than partial correlations.

Figure~\ref{fig:allocation_correlations} reports the resulting six
relationships, three candidate signals crossed with the two allocation behaviors, for every model and exam length. The correlations are first calculated per exam, then averaged over all exams with the same $N$.

% For token effort behavior, the correlations are calculated by pooling the question-level observations across all
% exams,
% Solving-order ranks, however, are meaningful only among questions that appear in the same exam and compete for the same budget. We therefore compute each solving-order correlation separately within an exam and then average the resulting correlations across exams.

\vspace{-0.05cm}

\paragraph{Models follow presentation order and increasingly neglect later questions.}
The position row of Figure~\ref{fig:allocation_correlations} reveals two
complementary behaviors. The blue effort cells indicate negative
correlations: questions presented later receive fewer reasoning tokens.
Averaged over all models and $N$s, this relationship is
$\rho=-0.34$, and it strengthens from $-0.17$ at $N=5$ to $-0.38$ at
$N=10$ and $-0.48$ at $N=20$. Thus, as more questions compete for the
same budget, reasoning becomes increasingly concentrated on the beginning
of the exam.

In the same position row, the red order cells indicate positive
correlations: the order in which models substantively work on questions
closely follows their presentation order. This relationship is strong
overall ($\rho=+0.68$) and remains stable across exam lengths
($+0.68$, $+0.66$, and $+0.69$). Models therefore largely follow the question presentation order regardless of the exam length, spending an increasingly large share on earlier questions as exams become longer.

\vspace{-0.05cm}

\paragraph{Difficulty affects effort reactively, not prospectively.}
Harder questions do receive more tokens, but the effect decays exactly as the budget grows tighter: $+0.33$ at $N{=}5$, $+0.26$ at $N{=}10$, and $+0.11$ at $N{=}20$. This is the signature of a reactive process. Once the model is inside a difficult question it keeps going, and at small $N$ it can afford to; it is not deciding in advance that a question deserves more compute, which would show up as a stable or strengthening relationship under pressure. Difficulty has little bearing on which question is taken up first: the order--difficulty correlation is $-0.03$ at $N{=}5$ and rises only to $+0.18$ at $N{=}20$. 
% That weak positive drift, a mild tendency to reach easy questions earlier on long exams, concentrated in the Qwen models, remains far below the position effect of $+0.69$ in the same cells.
\vspace{-0.05cm}

\paragraph{Point values have little effect.}
Neither allocation behavior responds meaningfully to the stated rewards.
Effort--value correlations are $0.00$, $+0.04$, and $+0.11$ across the
three exam lengths, while order--value correlations are also near zero ($-0.03$, $-0.09$, and $-0.07$). Even at $N=20$, where the
largest effort--value association appears, it is far less meaningful than the
corresponding position effect of $-0.48$.

\vspace{-0.05cm}

\paragraph{The pattern holds across model families.}
Models differ in how strongly their solving order follows presentation
position, but it remains the dominant signal for every
model. The association is strongest for DQ-7, DQ-14, and DSV4-F
($\rho=0.87$, $0.90$, and $0.91$), while the Qwen models depart from the
presented sequence more often ($0.55$, $0.45$, and $0.40$ for
QW-8/14/32). DSV4-P lies between these groups at $0.66$. Despite this
variation, all models show a negative effort--position correlation and
little sensitivity to point value. 

\vspace{-0.05cm}

\subsection{Is the Budget Spent on the Right Questions?}
\label{sec:right_questions}

Section~\ref{sec:allocation_results} showed that models organize their
reasoning largely by presentation position. This is harmful only if they
reach fewer questions than the budget allows, or if the questions they reach are not the
most worthwhile ones. We examine these two possibilities in turn.

\paragraph{Coverage shrinks as the exam grows.}
We measure substantive reach using the work set $W$ from
Equation~\ref{eq:work_set}. Table~\ref{tab:coverage} reports both its size
$|W|$ and the resulting coverage $|W|/N$. We additionally report the
\emph{zero-token rate}, the fraction of questions receiving no attributed
reasoning tokens at all. This is a more lenient notion than exclusion from
$W$: a question that is merely mentioned is not in the work set, but it
does not count as zero-token.

Across the locally deployed open models, the average work set grows from 4.0 questions at
$N=5$ to only 8.1 at $N=20$, so coverage falls from $80\%$ to $40\%$.
The API models are even more concentrated. DSV4-F works on 4.5 questions
at $N=10$ and 4.7 at $N=20$, reaching only $23\%$ of the longer exam.
At $N=20$, the zero-token rate is $51\%$ for the open-model average,
$69\%$ for DSV4-F, and $61\%$ for DSV4-P. Thus, as the exam grows, the
work set expands only slowly while an increasing fraction of questions
is never meaningfully considered.

\begin{table}[t]
    \centering
    \scriptsize
    \setlength{\tabcolsep}{2.7pt}
    \caption{
        Work-set size, coverage, and zero-token rate under fixed scoring,
        random order, and the baseline prompt. Columns within each group
        correspond to $N\in\{5,10,20\}$.
    }
    \label{tab:coverage}
    \resizebox{\columnwidth}{!}{%
    \begin{tabular}{@{}lccccccccc@{}}
        \toprule
        &
        \multicolumn{3}{c}{\textbf{Coverage} $\boldsymbol{|W|/N}$}
        &
        \multicolumn{3}{c}{\textbf{Work set} $\boldsymbol{|W|}$}
        &
        \multicolumn{3}{c}{\textbf{Zero-token rate}} \\
        \cmidrule(lr){2-4}
        \cmidrule(lr){5-7}
        \cmidrule(lr){8-10}
        \textbf{Model}
        & \textbf{5} & \textbf{10} & \textbf{20}
        & \textbf{5} & \textbf{10} & \textbf{20}
        & \textbf{5} & \textbf{10} & \textbf{20} \\
        \midrule
        DQ-7   & 0.55 & 0.49 & 0.38 & 2.8 & 4.9 & 7.6 & 0.34 & 0.48 & 0.54 \\
        DQ-14  & 0.90 & 0.72 & 0.46 & 4.5 & 7.2 & 9.2 & 0.08 & 0.23 & 0.46 \\
        QW-8   & 0.78 & 0.52 & 0.34 & 3.9 & 5.2 & 6.7 & 0.20 & 0.42 & 0.60 \\
        QW-14  & 0.90 & 0.67 & 0.40 & 4.5 & 6.7 & 8.0 & 0.08 & 0.25 & 0.49 \\
        QW-32  & 0.88 & 0.64 & 0.44 & 4.4 & 6.4 & 8.8 & 0.10 & 0.28 & 0.44 \\
        \midrule
        DSV4-F & 0.64 & 0.45 & 0.23 & 3.2 & 4.5 & 4.7 & 0.31 & 0.47 & 0.69 \\
        DSV4-P & 0.64 & 0.48 & 0.29 & 3.2 & 4.8 & 5.9 & 0.26 & 0.46 & 0.61 \\
        \bottomrule
    \end{tabular}%
    }
\end{table}

\paragraph{Viewing the exam as a knapsack with model-adaptive value density.}
Limited coverage does not yet establish that models reach the wrong
questions. To evaluate selection, we view the question selection problem in an exam as a knapsack problem~\cite{kellerer2004knapsack}: the
shared budget $B$ is the capacity, the point value $v_i$ is the value of
question $i$, and the tokens required to solve it are its weight $w_i$.
A natural greedy policy would prioritize questions with high value
density $\delta_i$.

As the shared-budget run cannot reveal $w_i$, we estimate it from an independent high-budget reference
condition, in which each question is solved independently with up to
40,960 tokens. For each model--question pair, the isolated token count
provides a model-adaptive estimation of $w_i$.
% while correctness indicates whether
% the model solved the question in this reference attempt.
We define
\begin{equation}
    \delta_i
    =
    \frac{v_i}{w_i}
    \,\mathbbm{1}
    \!\left[
        \hat{a}^{\,\mathrm{unc}}_i \text{ is correct}
    \right],
    \label{eq:density}
\end{equation}
where the indicator is 1 only when the model answers question $i$
correctly in the high-budget reference attempt. Questions that remain incorrect are
therefore assigned zero density.
% This construction is model-adaptive. The weight $w_i$ reflects the
% empirical cost of the question for the particular model being evaluated,
% rather than a single external difficulty label.
We do not interpret $w_i$ as the minimum or necessary cost of solving question $i$. It is the token usage observed in one independent high-budget attempt, which we use as a model-specific empirical difficulty proxy when computing value density.
% $\delta_i$ is intended to be model-adaptive as the same question may be
% cheap for one model, expensive for another, and unsolvable for a third.

\paragraph{Selection is blind to value density.}
For each exam, let $k=|W|$. We compare the work set with two reference
sets of the same size:

\begin{itemize}[leftmargin=12pt,itemsep=-2pt,topsep=1pt]
    \item the $k$ questions with the highest value density;
    \item the $k$ questions presented earliest in the prompt.
\end{itemize}
We call the overlapping ratio between $W$ and these sets as \emph{top-density overlap} and
\emph{early-position overlap}, respectively. A density-greedy policy would have
top-density overlap near 1, while the expected overlap by chance from selecting
$k$ questions without regard to either ranking is $k/N$.

Table~\ref{tab:density} shows that the work set is not more aligned with
value density than chance. Under random scoring and ascending difficulty
order, mean top-density overlap is $0.59$, indistinguishable from the
chance reference of $0.59$, whereas early-position overlap is $0.76$.
The same pattern holds under the other scoring schemes: top-density
overlap remains at or below chance, while early-position overlap stays
between $0.75$ and $0.81$.

Models also spend substantial compute on questions with $\delta_i=0$.
Such questions account for $24\%$ of the work set and $32\%$ of all
reasoning tokens on average, despite being answered incorrectly in the independent high-budget reference.

\begin{table}[t]
    \centering
    \scriptsize
    \setlength{\tabcolsep}{2.8pt}
    \caption{
        Work-set selection at $N{=}10$. The upper part uses random
        scoring and ascending difficulty order; the lower part reports
        means over models under the remaining scoring schemes
        (per-model breakdowns in Appendix~\ref{append:density}).
        Asterisks (*) indicate overlaps that are statistically significantly different from the "by chance" overlap based on the 95\% confidence interval.
    }
    \label{tab:density}
    \resizebox{\columnwidth}{!}{%
    \begin{tabular}{@{}lccccc@{}}
        \toprule
        \multirow{2}{*}{\textbf{Model}}
        & \multicolumn{3}{c}{
            \textbf{Set overlap ratio with $\boldsymbol{W}$}
        }
        & \multirow{2}{*}{
            \shortstack[c]{\textbf{Work set}\\[-1.5pt]
            $\boldsymbol{\delta=0}$}
        }
        & \multirow{2}{*}{
            \shortstack[c]{\textbf{Token share}\\[-1.5pt]
            $\boldsymbol{\delta=0}$}
        } \\
        \cmidrule(lr){2-4}
        &
        \textbf{By chance}
        &
        \textbf{Top-density}
        &
        \textbf{Early-position}
        &
        & \\
        \midrule
        DQ-7   & 0.62 & 0.62 & 0.86$^{*}$ & 0.39 & 0.51 \\
        DQ-14  & 0.63 & 0.65 & 0.75$^{*}$ & 0.39 & 0.48 \\
        QW-8   & 0.57 & 0.55 & 0.77$^{*}$ & 0.27 & 0.35 \\
        QW-14  & 0.65 & 0.67 & 0.76$^{*}$ & 0.20 & 0.26 \\
        QW-32  & 0.61 & 0.64 & 0.75$^{*}$ & 0.15 & 0.20 \\
        DSV4-F & 0.55 & 0.53 & 0.76$^{*}$ & 0.18 & 0.27 \\
        DSV4-P & 0.50 & 0.46$^{*}$ & 0.64$^{*}$ & 0.13 & 0.18 \\
        \midrule
        \textbf{Mean}
        & \textbf{0.59}
        & \textbf{0.59}
        & \textbf{0.76$^{*}$}
        & \textbf{0.24}
        & \textbf{0.32} \\
        \midrule
        \multicolumn{6}{@{}l}{
            \emph{Mean over models, other scoring schemes.}
        } \\
        Fixed    & 0.53 & 0.47$^{*}$ & 0.81$^{*}$ & 0.30 & 0.38 \\
        Aligned  & 0.51 & 0.46$^{*}$ & 0.78$^{*}$ & 0.31 & 0.39 \\
        Reversed & 0.54 & 0.53 & 0.75$^{*}$ & 0.28 & 0.35 \\
        \bottomrule
    \end{tabular}%
    }
\end{table}

Taken together, these results sharpen the position-driven failure from
\S~\ref{sec:allocation_results}. As exams grow, models reach only a
slowly expanding subset of questions. Within that subset, selection
matches the beginning of the prompt far better than the questions with
the highest model-adaptive value density, while a substantial fraction
of the budget is spent on questions the same model did not solve in
the independent high-budget reference.

% https://chatgpt.com/share/6a702ac5-b7ac-83e8-b802-6a036ed92821

\subsection{Does Prompting Improve Allocation?}
\label{sec:planning}

The base prompt states the budget and the goal but gives no guidance on
dividing it. We test four instructions that do: \emph{plan} the
allocation before solving, hint that questions may be \emph{skipped},
hint that answers should be \emph{rechecked}, and all three combined.
The details of each prompt are in Appendix~\ref{append:Prompts}.

\paragraph{Planning changes the allocation the most.}
Planning helps most, by a margin that widens as the budget tightens
(Table~\ref{tab:prompts}): averaged across five locally hosted open-weight models, coverage gains $0.09$, $0.12$, and $0.14$
over the base prompt as $N$ grows from $5$ to $20$. The skip hint also
improves coverage at every length, but by roughly half as much. The
recheck hint is flat or slightly harmful. Combining all three
reproduces the planning result rather than improving on it, so we
focus on planning below.

\begin{table}[t]
    \centering
    \small
    \caption{
        Effect of prompt instructions under fixed scoring and random
        order, averaged over the five locally hosted open-weight models. Within each
        group the three columns are $N{=}5$, $10$, and $20$. The best
        and second best are bold and underlined. Asterisks (*) mark a
        statistically significant difference from base prompt based on 95\% confidence interval.
    }
    \label{tab:prompts}

    % Small change value displayed at the lower right
    \newcommand{\change}[1]{%
        \hspace{0.05em}%
        \raisebox{-0.45ex}[0pt][0pt]{\tiny $(#1)$}%
    }

    \setlength{\tabcolsep}{4pt}
    \resizebox{\columnwidth}{!}{%
        \begin{tabular}{@{}lllllll@{}}
            \toprule
            & \multicolumn{3}{c}{\textbf{Work set coverage}}
            & \multicolumn{3}{c}{\textbf{Zero token rate}} \\
            \cmidrule(lr){2-4}
            \cmidrule(lr){5-7}
            \textbf{Prompt}
            & N=$\boldsymbol{5}$ & N=$\boldsymbol{10}$ & N=$\boldsymbol{20}$
            & N=$\boldsymbol{5}$ & N=$\boldsymbol{10}$ & N=$\boldsymbol{20}$ \\
            \midrule
            Base
            & 0.80
            & 0.61
            & 0.40
            & 0.16
            & 0.33
            & 0.51 \\

            Plan
            & \textbf{0.89$^{*}$}\change{+.09}
            & \underline{0.73$^{*}$}\change{+.12}
            & \textbf{0.54$^{*}$}\change{+.14}
            & \textbf{0.09$^{*}$}\change{-.07}
            & \underline{0.19$^{*}$}\change{-.14}
            & \textbf{0.32$^{*}$}\change{-.19} \\

            Skip hint
            & 0.86$^{*}$\change{+.06}
            & 0.69$^{*}$\change{+.08}
            & 0.45$^{*}$\change{+.05}
            & 0.11$^{*}$\change{-.05}
            & 0.24$^{*}$\change{-.09}
            & 0.43$^{*}$\change{-.08} \\

            Recheck hint
            & 0.79\change{-.01}
            & 0.57$^{*}$\change{-.04}
            & 0.37\change{-.03}
            & 0.18\change{+.02}
            & 0.38$^{*}$\change{+.05}
            & 0.54\change{+.03} \\

            \midrule
            All three
            & \textbf{0.89$^{*}$}\change{+.09}
            & \textbf{0.74$^{*}$}\change{+.13}
            & \underline{0.51$^{*}$}\change{+.11}
            & \textbf{0.09$^{*}$}\change{-.07}
            & \textbf{0.18$^{*}$}\change{-.15}
            & \underline{0.35$^{*}$}\change{-.16} \\
            \bottomrule
        \end{tabular}%
    }
\vspace{-0.3cm}
\end{table}

\paragraph{Planning spreads computation without redirecting it.}
Broken down by model (Table~\ref{tab:planning}), every open model except
QW-32 gains coverage, and DSV4-F coverage improves from $0.23$ to $0.33$. What
does not change is the basis on which questions are chosen. Solving
order stays tied to prompt position, with the open-model correlation
moving only from $0.64$ to $0.60$ and DSV4-P becoming \emph{more}
sequential ($0.68$ to $0.83$). Effort remains nearly uncorrelated to
point value.

\begin{table}[h]
    \centering
    \small
    \caption{
        Effect of explicit planning at $N{=}20$ by model. Coverage and
        zero rate use fixed scoring and random order; the value
        correlation uses the matched random-scoring condition.
        % Asterisks on Plan mark a difference from Base at $95\%$ by a
        % paired within-model exam bootstrap ($2000$ resamples).
         Asterisks (*) mark a
        statistically significant difference from base prompt based on 95\% confidence interval.
    }
    \label{tab:planning}
    \setlength{\tabcolsep}{3.5pt}
    \resizebox{\columnwidth}{!}{%
        \begin{tabular}{@{}lcccccccc@{}}
            \toprule
            & \multicolumn{2}{c}{\textbf{Coverage}}
            & \multicolumn{2}{c}{\textbf{Zero rate}}
            & \multicolumn{2}{c}{$\boldsymbol{\rho_{o,\pi}}$}
            & \multicolumn{2}{c}{$\boldsymbol{\rho_{t,v}}$} \\
            \cmidrule(lr){2-3}
            \cmidrule(lr){4-5}
            \cmidrule(lr){6-7}
            \cmidrule(lr){8-9}
            \textbf{Model}
            & \textbf{Base} & \textbf{Plan}
            & \textbf{Base} & \textbf{Plan}
            & \textbf{Base} & \textbf{Plan}
            & \textbf{Base} & \textbf{Plan} \\
            \midrule
            DQ-7   & 0.38 & 0.58$^{*}$ & 0.54 & 0.30$^{*}$ & 0.88 & 0.67$^{*}$ & $+0.09$ & $+0.05$ \\
            DQ-14  & 0.46 & 0.69$^{*}$ & 0.46 & 0.18$^{*}$ & 0.97 & 0.64$^{*}$ & $+0.11$ & $+0.10$ \\
            QW-8   & 0.34 & 0.46$^{*}$ & 0.60 & 0.37$^{*}$ & 0.54 & 0.74$^{*}$ & $+0.17$ & $+0.12$ \\
            QW-14  & 0.40 & 0.53$^{*}$ & 0.49 & 0.33$^{*}$ & 0.44 & 0.48 & $+0.23$ & $+0.09^{*}$ \\
            QW-32  & 0.44 & 0.43 & 0.44 & 0.41 & 0.39 & 0.45 & $+0.20$ & $+0.03^{*}$ \\
            \rowcolor{gray!15}
            mean   & 0.40 & 0.54 & 0.51 & 0.32 & 0.64 & 0.60 & $+0.16$ & $+0.08$ \\
            \midrule
            DSV4-F & 0.23 & 0.33$^{*}$ & 0.69 & 0.51$^{*}$ & 0.92 & 0.69$^{*}$ & $-0.03$ & $+0.01$ \\
            DSV4-P & 0.29 & 0.27 & 0.61 & 0.61 & 0.68 & 0.83 & $+0.00$ & $+0.04$ \\
            \rowcolor{gray!15}
            mean   & 0.26 & 0.30 & 0.65 & 0.56 & 0.80 & 0.76 & $-0.01$ & $+0.03$ \\
            \bottomrule
        \end{tabular}%
    }
    \vspace{-0.5cm}
\end{table}

Planning therefore changes the \emph{spread} of computation, not its
\emph{priorities}: instructed to budget its compute, the model divides
it more evenly rather than directing it anywhere in particular. This falls short of the objective because spreading tokens more uniformly is only beneficial when the questions that receive additional effort are actually worth solving.

% \vspace{-0.1cm}
\subsection{Does it Adapt to Position and Value?}
\label{sec:order_price}

\begin{figure}[h]
    \centering
    \includegraphics[width=0.8\columnwidth]{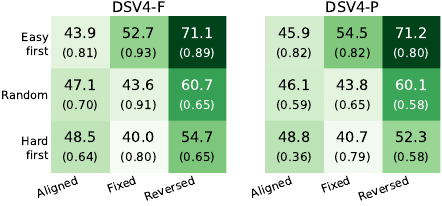}
    \caption{
        Score rate (\%, also shown by shading) and, in parentheses, the
        order--position correlation in the same setting, for DSV4-F and
        DSV4-P under the base prompt, averaged over
        $N\in\{5,10,20\}$. 
    }
    \label{fig:order_scoring}
\end{figure}

We next ask how allocation changes when the same questions are presented in a different order or under a different scoring scheme. Figure~\ref{fig:order_scoring} reports the score rate (Equation~\ref{eq:score_rate}) of the two strongest models, DSV4-F and DSV4-P; results for the remaining models are in Appendix~\ref{append:order-price}.
\vspace{-0.1cm}
\vspace{-0.2cm}

Under reversed scoring and easy-first presenting order, a position-driven sequential policy is near optimal because the earliest questions are both cheapest and most valuable. Under a hard-first presenting order, however, the models keep following the presenting position in prompt: correlations between solving order and position stay near $0.6$, and scores fall by $16$ to $19$ points on average. 
% Fixed scoring shows the same drop without a value confound, while aligned scoring, where the difficult questions presented first are also the most valuable, yields no compensating gain. 
This shows that the models are effectively hijacked by the most difficult questions up front, which also bear the lowest values.
Despite their strength, these models think hard through an adversarial sequence rather than reordering toward a smarter one.

\subsection{Results on Code Reasoning: CRUXEval-O}
\label{sec:crux_results}

We also run the shared-budget setting on CRUXEval-O~\citep{pmlr-v235-gu24c}, where the model predicts the return value of a short Python function. For each $N\in\{10,20\}$ we build 50 exams and evaluate QW-14, DQ-14, DSV4-F, and DSV4-P under fixed and random scoring, random order, and the base and explicit-planning prompts, with a pressure-matched budget $B=3{,}000$ (see Appendix~\ref{append:crux-budget} for the calibration). Since difficulty labels are unavailable in this dataset, we only use fixed scoring in the evaluation.

\begin{table}[h]
    \centering
    \scriptsize
    \setlength{\tabcolsep}{3.5pt}
    \caption{
        Allocation on CRUXEval-O under the base prompt and random
        question order.
    }
    \label{tab:crux_main}
    \resizebox{0.9\columnwidth}{!}{%
    \begin{tabular}{@{}lcccccc@{}}
        \toprule
        \textbf{Model}
        & $\boldsymbol{N}$
        & $\boldsymbol{\rho_{t,\pi}}$
        & $\boldsymbol{\rho_{o,\pi}}$
        & $\boldsymbol{\rho_{t,v}}$
        & \textbf{Coverage}
        & \textbf{Zero token rate} \\
        \midrule
        QW-14  & 10 & $-0.39$ & 1.00 & $+0.05$ & 0.49 & 24\% \\
        QW-14  & 20 & $-0.67$ & 1.00 & $+0.03$ & 0.21 & 49\% \\
        \midrule
        DQ-14  & 10 & $-0.16$ & 0.94 & $-0.06$ & 0.50 & 24\% \\
        DQ-14  & 20 & $-0.59$ & 1.00 & $-0.01$ & 0.22 & 44\% \\
        \midrule
        DSV4-F & 10 & $-0.20$ & 0.97 & $-0.06$ & 0.58 & 19\% \\
        DSV4-F & 20 & $-0.65$ & 1.00 & $+0.01$ & 0.23 & 40\% \\
        \midrule
        DSV4-P & 10 & $-0.13$ & 0.98 & $+0.15$ & 0.60 & 8\% \\
        DSV4-P & 20 & $-0.59$ & 1.00 & $+0.02$ & 0.23 & 30\% \\
        \bottomrule
    \end{tabular}%
    }
\end{table}

The same allocation pattern appears (Table~\ref{tab:crux_main}). Averaged across models and lengths, token effort declines with presentation position ($\rho=-0.42$) while solving order follows prompt position almost exactly ($\rho=0.99$), and effort remains nearly unrelated to stated point values ($\rho\approx0.02$). As $N$ doubles, coverage collapses to $0.21$--$0.23$ at $N{=}20$, with $30\%$--$49\%$ of questions receiving zero tokens. Further analysis are in Appendix~\ref{append:crux}.
% Explicit planning again fails to redirect priorities: at $N{=}20$, coverage stays within $0.04$ of the baseline for every model (Appendix~\ref{append:crux}).

% \input{sec/analysis}
\section{Conclusion}

% Reasoning models can spend test-time compute well on a single problem but cannot ration it across many: under a shared budget they solve in presentation order, front-load effort on early questions, ignore stated point values, and leave a growing fraction of the exam untouched as $N$ increases. The budget is consumed almost in full, but on questions selected by their position in the prompt rather than by their value or cost, and a substantial share goes to items the same model did not solve in the independent high-budget reference condition. Prompting models to plan first flattens spend without inducing strategic prioritization; score swings under alternate scorings reprice a largely fixed solved set rather than evidence of value-aware control. The same position-greedy, value-blind, $N$-worsening pattern appears on CRUXEval-O under a pressure-matched budget, with planning again flattening rather than strategizing. 
% Future work should test broader domains, train for cross-task allocation and metacognition, and build proper per-question cost oracles at matched budgets.

We introduce an exam-style framework for evaluating how reasoning models allocate a shared test-time compute budget across multiple questions. Our findings show that models consistently behave as position-driven sequential solvers: they prioritize questions in presentation order, concentrate effort on early items, respond weakly to point values, and leave increasingly many questions unattempted as the exam grows. Their selected questions align more closely with prompt position than with model-adaptive value density, and substantial compute is spent on problems unsolvable even under the high-budget reference condition. Explicit planning improves coverage by spreading effort more evenly, but does not produce value- or difficulty-aware prioritization. These results show that strong per-question reasoning does not imply effective global compute allocation, identifying shared-budget metareasoning as a distinct and unresolved capability for current reasoning models.
\section*{Limitations}
\label{sec:limitations}

\paragraph{Scope.}
Omni-MATH is the only domain in which we run the full factorial design. The CRUXEval-O experiments cover two exam lengths, four models, and two scoring schemes, with no native difficulty axis and no ordering or repricing manipulations. Absolute score rates and effect magnitudes should therefore not be compared across the two domains.

\paragraph{Token attribution.}
Per-question token counts in the shared-budget runs are recovered from \texttt{Q}$n$:\ marker segmentation, which is reliable for the large majority of traces but remains an approximation. Because a mention is not the same as an attempt, we report the work set and the token-centroid rank rather than raw mention counts throughout. A fully rigorous notion of effort, which would require detecting where a question is actually being solved rather than merely referenced, remains open.

\section*{Acknowledgments}

This work was supported in part by NSF CAREER Award 1942230, the ONR PECASE Award N00014-25-1-2378, ARO Early Career Program Award 310902-00001, Army Grant W911NF-21-2-0076, NSF Award CCF-2212458, NSF Award 2229885 (NSF Institute for Trustworthy AI in Law and Society, TRAILS), MURI Grant 14262683, DARPA AIQ Grant HR00112590066, and a Meta Research Award 314593-00001.

\bibliography{acl_latex}

\clearpage
\appendix
\section{Prompts and judge instructions}
\label{append:Prompts}

Figure~\ref{fig:inference_prompt} gives the shared-budget exam prompt, including the optional planning, skip, and recheck hints used in Section~\ref{sec:planning}. Figure~\ref{fig:judge_prompt} shows the LLM-as-judge instructions for mathematical answers; the deterministic CRUXEval-O matcher is described below.

\begin{figure*}[h]
\centering
\begin{tcolorbox}[
  enhanced,
  colframe=black!50,
  colback=white,
  coltitle=black,
  colbacktitle=black!20,
  width=\textwidth,
  arc=2mm,
  auto outer arc,
  boxrule=0.5pt,
  left=10pt,
  right=10pt,
  top=8pt,
  bottom=8pt,
  title=\textbf{Prompt to the LLM During Inference},
  fonttitle=\bfseries,
]
Below is a list of questions that you need to answer. Each question has an associated score (as shown in ``This question is worth ...'') and your goal is to maximize the total score obtained. For each question, you get full score if you answer it correctly, and zero if you answer it incorrectly. You in total have a reasoning budget of \emph{budget} tokens.

\medskip
\textbf{Optional strategy hints used in some conditions.} We additionally include one or more of the following sentences depending on the experimental setting:
\begin{itemize}[leftmargin=12pt]
    \item \emph{Explicit plan:} ``Each question has a different difficulty and may require a different amount of reasoning to answer correctly. You should plan your allocation/spending wisely to maximize the total score.''
    \item \emph{Skip hint:} ``You may even choose to not answer some questions if you think the cost of answering them is too high compared to the potential score gain.''
    \item \emph{Recheck hint:} ``If you think you have more than enough budget, you may also recheck and refine your answers to early questions to further increase the chances of maximizing the total score, but keep in mind that the rechecking also consumes your reasoning budget.''
\end{itemize}

\medskip
Questions:
\begin{verbatim}
Q1: [question text] (This question is worth [score] points)
Q2: [question text] (This question is worth [score] points)
...
QN: [question text] (This question is worth [score] points)
\end{verbatim}

Think through each question carefully to maximize your total score. Begin your reasoning now.
\end{tcolorbox}
\caption{The prompt used during inference. The exact wording varies slightly depending on the experimental condition through optional strategy hints.}
\label{fig:inference_prompt}
\end{figure*}
\begin{figure*}[h]
\centering
\begin{tcolorbox}[
  enhanced,
  colframe=black!50,
  colback=white,
  coltitle=black,
  colbacktitle=black!20,
  width=\linewidth,
  arc=2mm,
  auto outer arc,
  boxrule=0.5pt,
  left=10pt,
  right=10pt,
  top=8pt,
  bottom=8pt,
  title=\textbf{Instruction to the LLM Judge},
  fonttitle=\bfseries,
  % fontupper=\ttfamily
]
You are an expert mathematics judge. Your task is to evaluate whether the model's answers match the reference answers for a set of math questions. The reference answers are always correct, and the model's answers may be correct, incorrect, or incomplete.
\\

\{original\_questions\_text\}

\{ref\_text\}

\{model\_text\}

\{score\_text\}
\\

Please evaluate each answer for correctness. Consider the following:
1. Mathematical equivalence (e.g., 0.25 == 1/4, sqrt(3) == $\sqrt{3}$)
2. Different forms of the same answer (e.g., simplified vs expanded forms)
3. LaTeX formatting differences should not affect correctness
\\

For each question, award the full points if the answer is correct, and 0 points if incorrect.
\\

Provide your evaluation in the following JSON format:

\begin{verbatim}
{{
    "evaluations": [
        {{
            "question": 1,
            "correct": true/false,
            "score_awarded": <points>,
            "explanation": "<brief explanation>"
        }},
        ...
    ],
    "total_score": <sum of all awarded scores>,
    "max_score": <sum of all possible scores>
}}
\end{verbatim}

Respond with ONLY the JSON, no additional text.

\end{tcolorbox}
\caption{The instruction to \texttt{GPT-5} for it to serve as an LLM-as-a-judge.}
\label{fig:judge_prompt}
\end{figure*}

\subsection{CRUXEval-O exact-match judge}
\label{append:crux-judge}

CRUXEval-O answers are Python literals, so we use a deterministic judge rather than an LLM. The judge strips answer wrappers, parses predictions and references with \texttt{ast.literal\_eval}, and compares structures recursively. It tolerates three representation-only differences observed during manual error analysis: a dictionary body missing its outer braces, list--tuple interchange with identical elements, and integer--digit-string interchange only when their canonical decimal forms match (e.g., \texttt{89} vs.\ \texttt{'89'}). String case and internal whitespace remain exact. This avoids semantic fuzz while recovering formatting false negatives.

\section{Comparison with uniform allocation}
\label{append:uniform}

We compare the shared-budget run with a simple equal split. In the \emph{uniform per-question} condition, each question is solved independently with a budget of $B/N$ tokens. The answer depends on exam length (Table~\ref{tab:uniform}). At $N=5$, the shared-budget setting outperforms uniform allocation for four of seven models and gains $2.6$ score points on average. At $N=10$ the comparison is nearly even. At $N=20$, every model performs worse under the shared budget, with an average difference of $-5.0$ points.

At small $N$, front-loading can occasionally help by allowing the model to complete a few questions with high confidence, and the comparison becomes consistently unfavorable only as the exam grows and coverage collapses. This condition should also not be treated as an oracle, because independent prompts remove cross-question interference in addition to enforcing equal allocation. It is a reference for asking whether the model's emergent allocation provides a consistent advantage over a naive split, and the mechanism-level conclusions of Section~\ref{sec:results} do not depend on it.

\begin{table}[h]
    \centering
    \small
    \setlength{\tabcolsep}{7pt}
    \begin{tabular}{crr}
        \toprule
        $\boldsymbol{N}$
        & \textbf{Shared wins}
        & $\boldsymbol{\Delta}$ \\
        \midrule
        5  & 4/7 & $+2.6$ \\
        10 & 3/7 & $-0.4$ \\
        20 & 0/7 & $-5.0$ \\
        \bottomrule
    \end{tabular}
    \caption{
        Shared-budget performance relative to uniform per-question
        allocation under aligned scoring, random order, and the
        baseline prompt. $\Delta$ is the mean score-rate difference
        in percentage points: shared minus uniform.
    }
    \label{tab:uniform}
\end{table}

\section{Order-position correlations and score rates for locally deployed open models}
\label{append:order-price}

Figure~\ref{fig:order_scoring} focuses on DSV4-F and DSV4-P. Table~\ref{tab:order_price_open} reports the same order$\times$scoring grid for the five open models, and Table~\ref{tab:order_price_n} separates the reversed-scoring hard-first penalty by exam length for all seven models. Easy-first presentation is already near a good sequential policy under reversed scoring, so the hard-first change measures escape from an adversarial order. The API models lose $11$--$22$ points at every length while retaining order--position correlations near $0.6$. Open models mostly match their easy-first scores; QW-32's correlation falls to $-0.25$ while its score does not drop, but that departure does not yield gains under aligned scoring either. DQ-7 retains the sequence ($\rho=0.71$) yet shows no penalty, scoring in the low teens under both orders.

\begin{table}[h]
    \centering
    \scriptsize
    \setlength{\tabcolsep}{3.5pt}
    \caption{
        Score rate (\%) and order--position correlation (in parentheses)
        for the open models under the baseline prompt, averaged over
        $N\in\{5,10,20\}$. Format matches Figure~\ref{fig:order_scoring}.
    }
    \label{tab:order_price_open}
    \resizebox{\columnwidth}{!}{%
    \begin{tabular}{@{}llccc@{}}
        \toprule
        \textbf{Model} & \textbf{Order}
        & \textbf{Aligned} & \textbf{Fixed} & \textbf{Reversed} \\
        \midrule
        \multirow{3}{*}{DQ-7}
        & Easy first & 10.2 ($0.74$) & 13.9 ($0.82$) & 15.9 ($0.76$) \\
        & Random     & 13.4 ($0.73$) & 15.5 ($0.86$) & 18.7 ($0.75$) \\
        & Hard first & 12.3 ($0.75$) & 15.4 ($0.85$) & 18.5 ($0.71$) \\
        \midrule
        \multirow{3}{*}{DQ-14}
        & Easy first & 23.8 ($0.69$) & 27.8 ($0.87$) & 39.6 ($0.86$) \\
        & Random     & 27.5 ($0.66$) & 27.7 ($0.90$) & 38.1 ($0.65$) \\
        & Hard first & 25.4 ($0.82$) & 27.5 ($0.90$) & 37.0 ($0.34$) \\
        \midrule
        \multirow{3}{*}{QW-8}
        & Easy first & 13.9 ($0.84$) & 16.3 ($0.90$) & 24.8 ($0.88$) \\
        & Random     & 15.1 ($0.52$) & 18.2 ($0.55$) & 23.9 ($0.42$) \\
        & Hard first & 13.8 ($0.42$) & 19.6 ($0.50$) & 25.4 ($0.22$) \\
        \midrule
        \multirow{3}{*}{QW-14}
        & Easy first & 21.2 ($0.67$) & 23.4 ($0.85$) & 36.7 ($0.80$) \\
        & Random     & 20.5 ($0.41$) & 25.2 ($0.45$) & 35.2 ($0.30$) \\
        & Hard first & 17.4 ($0.40$) & 24.7 ($0.34$) & 35.0 ($0.12$) \\
        \midrule
        \multirow{3}{*}{QW-32}
        & Easy first & 24.5 ($0.64$) & 28.4 ($0.80$) & 38.7 ($0.73$) \\
        & Random     & 22.3 ($0.39$) & 27.6 ($0.40$) & 41.9 ($0.24$) \\
        & Hard first & 22.9 ($0.55$) & 28.8 ($0.26$) & 40.7 ($-0.25$) \\
        \bottomrule
    \end{tabular}%
    }
\end{table}

\begin{table}[h]
    \centering
    \small
    \setlength{\tabcolsep}{7pt}
        \caption{
        Score-rate change (percentage points) from hard-first relative
        to easy-first presentation under reversed scoring and the
        baseline prompt. Negative values indicate a loss from the
        adversarial order.
    }
    \label{tab:order_price_n}
    \begin{tabular}{lrrr}
        \toprule
        \textbf{Model}
        & $\boldsymbol{N{=}5}$
        & $\boldsymbol{10}$
        & $\boldsymbol{20}$ \\
        \midrule
        DSV4-P & $-21.9$ & $-17.7$ & $-17.1$ \\
        DSV4-F & $-11.5$ & $-22.1$ & $-15.7$ \\
        \midrule
        DQ-14  & $+4.5$ & $-7.9$ & $-4.5$ \\
        QW-14  & $-5.9$ & $+0.3$ & $+0.4$ \\
        QW-8   & $-1.0$ & $+1.1$ & $+1.7$ \\
        QW-32  & $+1.8$ & $+2.7$ & $+1.7$ \\
        DQ-7   & $+4.3$ & $+1.4$ & $+2.0$ \\
        \bottomrule
    \end{tabular}
\end{table}

\section{CRUXEval-O details}
\label{append:crux}

\subsection{Budget calibration}
\label{append:crux-budget}

To make the constraint comparable with the mathematics setting, we estimate a typical independent high-budget solution cost from the token usage of correctly answered items and approximately match
\begin{equation}
    \frac{B}{
        N \times
        \text{average reference cost per question}
    }
\end{equation}
across domains. The median high-budget reference cost for CRUXEval-O is approximately 995 tokens per correctly answered question, giving $B=3{,}000$ as roughly three times that median.

\subsection{Budget exhaustion and planning}

Runs almost always spend the entire budget, and increasingly so with exam length: the fraction of runs that reach the $3{,}000$-token limit rises from $56\%$--$88\%$ at $N=10$ to $96\%$--$100\%$ at $N=20$.

Table~\ref{tab:crux_plan} reports explicit planning at $N{=}20$. Coverage stays within $0.04$ of the baseline for every model, and effort--value correlations remain weak under both prompts. This is a weaker coverage response than in mathematics (Section~\ref{sec:planning}), where planning raised open-model coverage by $0.14$ at $N{=}20$: with a budget of roughly three times the cost of a single question, spreading it more evenly is largely unavailable even when instructed.

\begin{table}[h]
    \centering
    \small
    \setlength{\tabcolsep}{5pt}
    \caption{
        Explicit planning on CRUXEval-O at $N{=}20$ under fixed scoring
        and random order. Value correlations use the matched
        random-scoring condition.
    }
    \label{tab:crux_plan}
    \begin{tabular}{@{}lcccc@{}}
        \toprule
        & \multicolumn{2}{c}{\textbf{Coverage}}
        & \multicolumn{2}{c}{$\boldsymbol{\rho(\text{eff.},\text{val.})}$} \\
        \cmidrule(lr){2-3}
        \cmidrule(lr){4-5}
        \textbf{Model}
        & \textbf{Base} & \textbf{Plan}
        & \textbf{Base} & \textbf{Plan} \\
        \midrule
        QW-14  & 0.21 & 0.20 & $+0.03$ & $+0.05$ \\
        DQ-14  & 0.22 & 0.26 & $+0.02$ & $+0.15$ \\
        DSV4-F & 0.23 & 0.21 & $+0.02$ & $-0.01$ \\
        DSV4-P & 0.23 & 0.26 & $+0.02$ & $+0.01$ \\
        \bottomrule
    \end{tabular}
\end{table}

\section{Work set selection under other scoring schemes}
\label{append:density}

Table~\ref{tab:density} reports per-model results under random scoring and model means under the remaining schemes. Table~\ref{tab:density_other} gives the per-model breakdown for fixed, aligned, and reversed scoring at $N{=}10$ (random order, baseline prompt). Top-density overlap remains at or below chance in every block, while early-position overlap stays well above chance.

\begin{table}[h]
    \centering
    \scriptsize
    \setlength{\tabcolsep}{3pt}
    \caption{
        Work-set selection at $N{=}10$ under fixed, aligned, and reversed
        scoring (random order, base prompt). Asterisks (*) indicate overlaps
        that are statistically significantly different from the ``by chance''
        overlap based on the 95\% confidence interval.
    }
    \label{tab:density_other}
    \resizebox{\columnwidth}{!}{%
    \begin{tabular}{@{}llccccc@{}}
        \toprule
        \multirow{2}{*}{\textbf{Scoring}}
        & \multirow{2}{*}{\textbf{Model}}
        & \multicolumn{3}{c}{
            \textbf{Set overlap ratio with $\boldsymbol{W}$}
        }
        & \multirow{2}{*}{
            \shortstack[c]{\textbf{Work set}\\[-1.5pt]
            $\boldsymbol{\delta=0}$}
        }
        & \multirow{2}{*}{
            \shortstack[c]{\textbf{Token share}\\[-1.5pt]
            $\boldsymbol{\delta=0}$}
        } \\
        \cmidrule(lr){3-5}
        &
        &
        \textbf{By chance}
        &
        \textbf{Top-density}
        &
        \textbf{Early-position}
        &
        & \\
        \midrule

        \multirow{8}{*}{Fixed}
        & DQ-7   & 0.55 & 0.52 & 0.88$^{*}$ & 0.50 & 0.59 \\
        & DQ-14  & 0.66 & 0.59$^{*}$ & 0.87$^{*}$ & 0.43 & 0.57 \\
        & QW-8   & 0.50 & 0.48 & 0.76$^{*}$ & 0.33 & 0.45 \\
        & QW-14  & 0.62 & 0.56$^{*}$ & 0.84$^{*}$ & 0.26 & 0.32 \\
        & QW-32  & 0.58 & 0.50$^{*}$ & 0.71$^{*}$ & 0.21 & 0.27 \\
        & DSV4-F & 0.43 & 0.35$^{*}$ & 0.85$^{*}$ & 0.20 & 0.24 \\
        & DSV4-P & 0.40 & 0.27$^{*}$ & 0.76$^{*}$ & 0.16 & 0.25 \\
        \rowcolor{gray!15}
        & Mean   & 0.53 & 0.47$^{*}$ & 0.81$^{*}$ & 0.30 & 0.38 \\
        \midrule

        \multirow{8}{*}{Aligned}
        & DQ-7   & 0.49 & 0.44 & 0.90$^{*}$ & 0.46 & 0.57 \\
        & DQ-14  & 0.56 & 0.52 & 0.82$^{*}$ & 0.49 & 0.61 \\
        & QW-8   & 0.49 & 0.45 & 0.78$^{*}$ & 0.32 & 0.45 \\
        & QW-14  & 0.61 & 0.56$^{*}$ & 0.80$^{*}$ & 0.25 & 0.31 \\
        & QW-32  & 0.49 & 0.47 & 0.57 & 0.20 & 0.26 \\
        & DSV4-F & 0.48 & 0.40$^{*}$ & 0.89$^{*}$ & 0.23 & 0.28 \\
        & DSV4-P & 0.44 & 0.41 & 0.73$^{*}$ & 0.18 & 0.23 \\
        \rowcolor{gray!15}
        & Mean   & 0.51 & 0.46$^{*}$ & 0.78$^{*}$ & 0.31 & 0.39 \\
        \midrule

        \multirow{8}{*}{Reversed}
        & DQ-7   & 0.50 & 0.51 & 0.85$^{*}$ & 0.40 & 0.46 \\
        & DQ-14  & 0.55 & 0.51 & 0.75$^{*}$ & 0.46 & 0.58 \\
        & QW-8   & 0.55 & 0.60 & 0.71$^{*}$ & 0.25 & 0.36 \\
        & QW-14  & 0.64 & 0.66 & 0.70 & 0.21 & 0.27 \\
        & QW-32  & 0.62 & 0.66 & 0.64 & 0.18 & 0.20 \\
        & DSV4-F & 0.46 & 0.35$^{*}$ & 0.84$^{*}$ & 0.23 & 0.26 \\
        & DSV4-P & 0.44 & 0.41 & 0.73$^{*}$ & 0.20 & 0.30 \\
        \rowcolor{gray!15}
        & Mean   & 0.54 & 0.53 & 0.75$^{*}$ & 0.28 & 0.35 \\
        \bottomrule
    \end{tabular}%
    }
\end{table}

\section{Token attribution details}
\label{append:attribution}

Per-question tokens are attributed by strict \texttt{Q}$n$:\ markers in the Phase~1 trace. As mention $\neq$ work: opening enumeration passes can name every question with minimal token mass. We therefore use (i) \emph{zero-token} rate (no attributed tokens), (ii) \emph{work set} ($\geq$200 tokens or $\geq$2 segments), and (iii) token-centroid rank \emph{within the work set} for solving order. First-mention order inflates sequentiality for DeepSeek models, which centroid rank corrects; the work-set restriction additionally prevents a question that is only named in an enumeration pass---whose centroid sits at that early offset---from being ranked ahead of questions that received real effort. The restriction matters at the head of the ranking---with all mentioned questions, the earliest-centroid question is a mention-only question in up to 56\% of traces---but barely moves the aggregate correlations: ranking all mentioned questions instead gives order$\sim$position $0.70$ (vs.\ $0.69$), order$\sim$difficulty $0.11$ (vs.\ $0.07$), and order$\sim$points $-0.03$ (vs.\ $-0.09$).

\end{document}